\documentclass[letterpaper, 10 pt, conference]{ieeeconf}

\IEEEoverridecommandlockouts                              %
\usepackage[T1]{fontenc}
\usepackage[utf8]{inputenc}

\usepackage{amsmath} %
\usepackage{amssymb} %

\usepackage{hyperref}
\usepackage{graphicx}
\usepackage{tabularx}
\usepackage{xcolor}
\definecolor{mygreen}{RGB}{80, 174, 95}
\definecolor{myred}{RGB}{184, 44, 44}

\usepackage{multirow}
\usepackage{booktabs}
\usepackage{placeins}

\usepackage{cleveref}

\usepackage{bm}
\usepackage{url}

\usepackage{cite}

\usepackage{ifthen}

\title{\LARGE \bf Multi-viewpoint Geo-localization with Event Cameras}

\author{Adam D. Hines$^{*}$  \quad Michael Milford \quad Tobias Fischer %
\thanks{The authors are with the QUT Centre for Robotics, Queensland University of Technology, Brisbane QLD 4001. $^{*}$Correspondence to {\tt adam.hines@qut.edu.au}.}%
\thanks{This work received funding from an ARC Laureate Fellowship FL210100156 to MM and an ARC Discovery Early Career Researcher Award DE240100149 to TF. The authors acknowledge continued support from the Queensland University of Technology (QUT) through the Centre for Robotics.}
 }%

\begin{document}
\bstctlcite{IEEEexample:BSTcontrol}
\maketitle
\thispagestyle{empty}
\pagestyle{empty}

\begin{abstract}%
\label{sec:abstract}%
Robot localization is an ongoing challenge that demands mapping and positioning systems that are tolerant to viewpoint change. Event cameras are attracting increasing interest and adoption in robotics; however, dealing with viewpoint variance is an under-investigated problem in existing event-based localizers. In addition, event-based datasets that emphasize viewpoint variance for challenging localization situations are scarce. Here, we introduce an event-based visual place recognition (VPR) system that performs robustly under viewpoint changes. We converted five large-scale geo-tagged datasets, conventionally used to train frame-based localization systems, into synthetic event streams using Image-to-Event (I2E) conversion, and used them to fine-tune a pre-trained event-based vision transformer backbone with a multi-loss function, yielding a system we call MegaEvent that learns viewpoint-robust features for place recognition. We achieved an average Recall@1 of 82\% across three existing event-based localization datasets, leading the next best event-based method by 20 recall points, and frame-based VPR models applied directly to event frames by 8 to 26 recall points. We introduce a new, challenging dataset -- \emph{Springfield-Event-VPR} -- which features a 3.7~km walking route recorded in three camera orientations for a total of 11.1~km, which MegaEvent outperforms the strongest baseline by 9 recall points. The code for MegaEvent is available at \url{https://github.com/AdamDHines/megaevent}.
\end{abstract}
\section{Introduction}
\label{sec:intro}
Event cameras respond at the pixel level to changes in light intensity~\cite{Gallego2022}, offering low-latency and low-power sensing for robotic perception. However, relatively few systems have demonstrated these advantages in real-world robotic autonomy~\cite{Falanga2020, Gallego2022, ParedesValles2024, Hines2025}. For event-based visual place recognition (VPR), a particular limitation is robustness to changes in viewpoint and camera rotation~\cite{Lee2021, Fischer2022, Hines2025, Keime2026}. In this work, we address this problem by developing an event-based VPR system for large-scale, viewpoint-robust localization.

    \begin{figure}[!t]
        \centering
        \includegraphics[width=\linewidth]{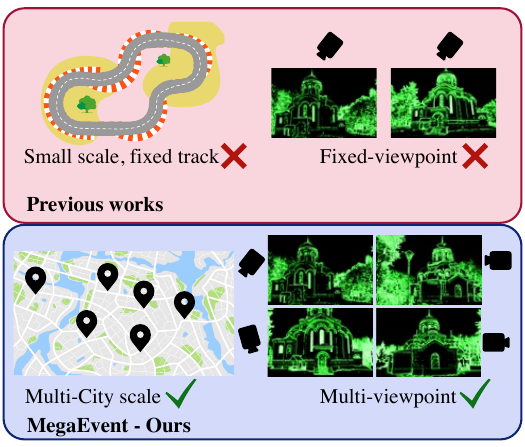}
        \caption{Currently available localization systems for event cameras constrain performance to small-scale, fixed-track areas and are view-variant. MegaEvent provides multi-viewpoint geo-localization capabilities, trained at multi-city scale.}
        \vspace{-15pt}
        \label{fig:schema}
    \end{figure}

Modern frame-based VPR systems achieve strong viewpoint robustness by training on large and diverse datasets containing millions of images captured across different viewpoints, locations, lighting conditions, and times of year~\cite{Warburg2020, Berton2022, Berton2025}. Equivalent training data are not available for event-based VPR. Existing event-based VPR datasets~\cite{Zhu2018, Fischer2020, Hu2020, Carmichael2025, Pan2025} provide important evaluation benchmarks, but are small compared with datasets such as SF-XL, which contains more than four million images~\cite{Berton2022}. This limits the ability to apply the large-scale training regimes that have driven recent advances in conventional VPR.

We therefore present MegaEvent, an event-based VPR model for viewpoint-robust, large-scale geo-localization (Fig.~\ref{fig:schema}). MegaEvent adapts the multi-dataset training regime of MegaLoc~\cite{Berton2025} to the event domain, exploiting recent developments in Image-to-Event (I2E) systems~\cite{Ma2026}. Following~\cite{Berton2025}, we converted five established VPR training datasets~\cite{Dai2017, Warburg2020, Ali2022, Berton2022, Tung2024} into synthetic event streams~\cite{Ma2026} for training MegaEvent. This produces more than 8 million geo-tagged event frames spanning multiple cities, viewpoints, lighting conditions, and environments. These data are used to fine-tune a pre-trained vision transformer (ViT) backbone~\cite{Cao2026} with a Sinkhorn Algorithm for Locally Aggregated Descriptors (SALAD) aggregator~\cite{Izquierdo2024} for event-based VPR. We additionally introduce a challenging multi-viewpoint event-based dataset for evaluation -- Springfield-Event-VPR.

Specifically, we contribute:
\begin{enumerate}
    \item A viewpoint-robust event-based place recognition system trained entirely on conventional VPR datasets converted to synthetic event streams.
    \item A new multi-viewpoint localization dataset -- called \emph{Springfield-Event-VPR} -- recorded on a Prophesee EVK4 event camera, comprising a 3.7~km walking route traversed in three camera orientations for a total of 11.1~km, and a set of queries captured under different lighting and extreme viewpoints.
    \item An analysis of how a model trained on synthetic event streams behaves when the same places are presented as real versus synthetically converted event streams.
\end{enumerate}

The code for MegaEvent is available at~\url{https://github.com/AdamDHines/megaevent}. The Springfield-Event-VPR dataset, with instructions and full GPS annotation, is available for evaluation at~\url{https://huggingface.co/datasets/AdamHines/springfield-event-vpr}.
\section{Related Works}
\label{sec:relatedworks}
Here, we survey recent advances in performing event-based VPR in Sect.~\ref{subsec:evpr}, vision backbones for event cameras in Sect.~\ref{subsec:dlevcam}, and methods for converting frame images from datasets to event streams in Sect.~\ref{subsec:evstreams}.

\subsection{Event-based visual place recognition}
\label{subsec:evpr}
Event-based VPR spans a broad range of applications and implementations. EventVLAD~\cite{Lee2021} introduces one of the first deep-learned methods, contributing a denoising network for event frames to extract NetVLAD features. Ensemble-Event-VPR~\cite{Fischer2020} uses the events-to-video (E2VID) system~\cite{Rebecq2019} to reconstruct intensity images from events alongside event frames, improving localization accuracy. Another ensembling method combines multiple state-of-the-art RGB systems to perform highly accurate localization~\cite{Joseph2025}.

Sparse-Event-VPR~\cite{Fischer2022} investigates the minimum number of event pixels required to perform accurate localization. Locational Encoding with Neuromorphic Systems~\cite{Hines2025} introduces a fully asynchronous localization system deployed on an energy-efficient neuromorphic processor. EventGeM~\cite{Hines2026} combines global and local feature detection from a pre-trained keypoint detector with RANSAC re-ranking for multi-feature VPR. Spike-EVPR~\cite{Liu2026} and SpikeVPR~\cite{Keime2026} apply spiking neural networks to event-based localization by generating spiking place descriptors.

Event-VPR~\cite{Kong2022} is the most similar method to our proposed MegaEvent. Whereas Event-VPR uses events generated from videos, we generate event streams from static images used in large-scale geo-localization datasets, including SF-XL and MegaScenes~\cite{Berton2022, Tung2024}.

\subsection{Vision backbones for event cameras}
\label{subsec:dlevcam}
Several vision backbones support processing event-based data~\cite{Yang2023, Yang2024, Cao2026}. Typically, these are trained using a student-teacher regime with RGB image outputs from vision foundation models. A limitation and by-product of training event-based vision backbones like this is that the event representation used is fixed. For example, a model trained on polarity histogram event frames would not be compatible with time surface or voxel based representations. Deploying several such systems on one platform therefore requires maintaining a separate event representation for each, duplicating the event pre-processing pipeline. Event-camera data pre-training~\cite{Yang2023} uses the N-ImageNet dataset with contrastive learning for object recognition, optical flow estimation, and semantic segmentation. Event-camera data dense pre-training~\cite{Yang2024} uses the Video-to-Events (V2E) system~\cite{Gehrig2020} to perform dense prediction tasks, such as depth estimation. More recently, generative event pre-training uses foundation model alignment between event streams and videos from several event datasets for state-of-the-art vision backbone performance~\cite{Cao2026}.

In this work, we use the current state-of-the-art backbone from~\cite{Cao2026} as the basis for MegaEvent, and consider both the small ($\approx$20M parameters) and base ($\approx$80M parameters) models. These figures count the backbone only; the totals reported in Table~\ref{tab:modelparams} additionally include the SALAD aggregator and the linear projection head.

\subsection{Generating event datasets from frame-based images}
\label{subsec:evstreams}
Many datasets exist for training frame-based models, which is not the case for event cameras. Conventional training datasets such as ImageNet~\cite{Deng2009} and COCO~\cite{Lin2014} owe much of their value to the dense human annotation they provide. Several methods now convert images and videos into event streams to train neuromorphic deep learning models. This is achieved in two main ways: recording events with a camera pointed at images presented on a screen, or synthetically deriving event streams.

Datasets captured with event cameras include N-MNIST~\cite{Orchard2015}, N-ImageNet~\cite{Kim2021}, N-Caltech101~\cite{Orchard2015}, and CIFAR10-DVS~\cite{Li2017}. For the student-teacher based vision backbones in Sect.~\ref{subsec:dlevcam}, N-ImageNet remains a common standard for general feature extraction and training for its breadth of images and rich annotation~\cite{Kim2021}.

Conversely, recent methods produce event streams directly from frame-based videos and, more recently, static images. ESIM~\cite{Rebecq2018} is among the first simulators to synthesize events from rendered scenes with adaptive temporal sampling, and V2E~\cite{Gehrig2020} extends this to generate realistic events from ordinary video sequences. Relevant to our work, I2E~\cite{Ma2026} generates event streams from static images using saccadic scanning patterns. Unlike prior work, which uses these simulators to produce pre-training data for low-level tasks, we convert entire geo-tagged VPR training sets, preserving the place labels that supervise retrieval.
\section{Methodology}
\label{sec:method}
We adapt the training regime of MegaLoc~\cite{Berton2025} to fine-tune the generative event backbone of Cao~et al.~\cite{Cao2026} for accurate VPR. This required the conversion of five individual large-scale geo-tagged localization datasets from static images to event streams, outlined in Sect.~\ref{subsec:datasets}. For the models in~\cite{Cao2026}, we fine-tuned based on a combined loss function described in Sect.~\ref{subsec:finetune}. We additionally designed noise parameters to simulate the dynamics of an event camera for images converted using I2E, to avoid over-fitting to a synthetic domain (Sect.~\ref{subsec:trainhyper}).

\subsection{Synthetic dataset generation}
\label{subsec:datasets}
To create a training regime that presents several geographically distinct places each with multiple viewpoints to learn viewpoint robustness, we follow the data organization of MegaLoc~\cite{Berton2025} for each of the five training datasets. Briefly, SF-XL~\cite{Berton2022} follows the sampling technique of EigenPlaces~\cite{Berton2023} to provide images of a place with multiple distinct views, split into two sub-batches of frontal and lateral facing images, and preventing visual overlap between different places. Google Street View Cities~\cite{Ali2022} consists of 40 unique cities with at least 100~m separation of places. MSLS~\cite{Warburg2020} is a long-term dataset captured over nine years, with place sampling following CliqueMining~\cite{Izquierdo2024-ECCV} retaining 100~m separation across places. MegaScenes~\cite{Tung2024} is a structure-from-motion reconstruction dataset, where each reconstruction is considered a class. ScanNet~\cite{Dai2017} consists of 707 unique indoor spaces, and we treat each scene as a training class while retaining some visual overlap between images of the same scene. Across all datasets, a quadruplet of images is selected from $N$ different places, following the image selection protocol from~\cite{Berton2025}.

\begin{figure}[t]
        \centering
        \includegraphics[width=\columnwidth]{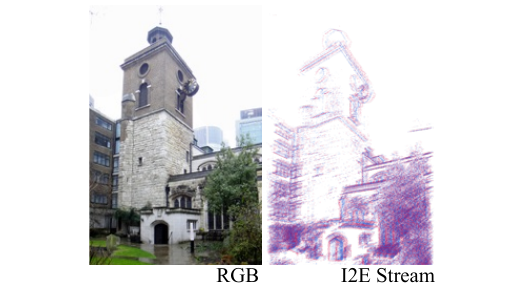}
        \caption{Example RGB image (left) and the corresponding raw synthetic Image-to-Event (I2E) stream (right), sampled from the MegaScenes dataset~\cite{Tung2024}. The stream is visualized directly; it is not the three-channel representation of~Eq.\eqref{eq:accumulate} that the backbone consumes.}
        \vspace{-15pt}
        \label{fig:i2e}
    \end{figure}

The event stream conversion from the conventional static images was performed using I2E~\cite{Ma2026}. I2E converts images to event streams by creating intensity maps from RGB images, simulating micro-saccadic eye movements through spatio-temporal convolution, and applying an adaptive event firing normalization stage. We set a maximum saccade time of $30$~ms for all synthetic event stream generation. Fig.~\ref{fig:i2e} shows an example of a synthetic event stream.

I2E produces a standard event stream consisting of events $\mathbf{e} = [x, y, t, p]$, where $x$ and $y$ are pixel coordinates, $t$ is the timestamp, and $p$ is the polarity (ON or OFF).

Event streams are stored and converted to the various event representations for each of the baseline methods implemented (Sect.~\ref{subsec:finetune} and Sect.~\ref{subsec:baselines}) with EventCV~\cite{eventcv}.

\subsection{Backbone model and aggregator fine-tuning}
\label{subsec:finetune}
The pre-trained backbone used for MegaEvent is the generative pre-trained backbone from~\cite{Cao2026}, which contributed both a small (ViT/S) and a base (ViT/B) model of $\approx$20M and $\approx$80M parameters, respectively. The main backbone is DINOv2~\cite{Oquab2024}, which is trained on a 3-channel RGB
event image representation, with positive and negative polarities rendered as red
and blue intensities on a white background and only the dominant polarity kept per
pixel. The accumulated count of each polarity is first normalized in the following
way:

    \begin{equation}
        X_e(x, y, c) = \frac{\min(M_c(x,y),\alpha_{q,c})}{\alpha_{q,c}}, \quad c \in \{p,n\},
    \label{eq:accumulate}
    \end{equation}
    where $M_c(x,y)$ is the raw accumulated count in polarity channel $c$, and $\alpha_{q,c}$ is a normalization scale taken as the $q$-th percentile ($q=99$) of the non-zero pixel values of each polarity channel separately~\cite{Cao2026}. The two normalized polarity magnitudes $X_e(x,y,p)$ and $X_e(x,y,n)$ are then mapped onto the three channels of the RGB image the backbone consumes: a positive event suppresses the green and blue channels so the pixel renders red, a negative event suppresses the red and green channels so the pixel renders blue, and a pixel with no events remains white. For real event streams, each frame is built from a 50~ms window of events, resized to $224\times224$~px for training, and $322\times322$~px for inference, following~\cite{Berton2025}.

We use the first 12 layers of the pre-trained backbone, prior to any task specific trained head, to fine-tune the model and the SALAD~\cite{Izquierdo2024} descriptor aggregator. Following~\cite{Berton2025}, each training step draws one sub-batch from each of six streams, with SF-XL frontal and lateral views treated separately. The multi-similarity loss~\cite{Wang2019} is computed independently within each sub-batch $\mathcal{B}_i$, such that positive and negative mining is restricted to samples from the same stream. The total loss is the equally weighted sum

    \begin{equation}
    \mathcal{L} = \sum_{i=1}^{6} \mathcal{L}_{\mathrm{MS}}(\mathcal{B}_i).
    \label{eq:losstot}
    \end{equation}

For fine-tuning, both ViT/S and ViT/B had all 12 layers unfrozen for training. The datasets, baselines, and hyperparameters used to train and evaluate this model are described next.
\section{Implementation \& Experimental Setup}
\label{sec:expsetup}

Here, we describe the datasets used for evaluation in Sect.~\ref{subsec:evaldata}, details on the new viewpoint variant Springfield-Event-VPR dataset contributed in this work (Sect.~\ref{subsec:noveldata_setup}), baseline comparison methods in Sect.~\ref{subsec:baselines}, the metric used to evaluate localization performance in Sect.~\ref{subsec:localmetrics}, training hyperparameters in Sect.~\ref{subsec:trainhyper}, and finally the details on the computational platform in Sect.~\ref{subsec:compenv}.

\subsection{Evaluation datasets}
\label{subsec:evaldata}
Localization performance was evaluated on three event-based datasets, specifically \textbf{Brisbane-Event-VPR~\cite{Fischer2020}}, \textbf{NSAVP~\cite{Carmichael2025}}, and \textbf{NYC-Event~\cite{Pan2025}}, as well as our newly introduced dataset.

All datasets had a localization tolerance set to 25~m, \emph{i.e.},~a match is deemed correct if the retrieved reference is within this tolerance from the true location, following~\cite{Berton2025}. Stops were not removed from any dataset. For NYC-Event-VPR we followed the dataset's own benchmark protocol~\cite{Pan2025}, holding out a random 10\% of places as queries against the remaining 90\% as the reference.

\subsection{Novel event VPR dataset}
\label{subsec:noveldata_setup}
The benchmark datasets evaluated so far are all captured ``on-the-rails'', with query and database traverses following the same route and heading, which makes evaluating viewpoint and rotation variance not possible. To that end, we contribute a new dataset called \textbf{\emph{Springfield-Event-VPR}} that, for the first time in event-based data, consists of challenging viewpoint shifts and heading changes relative to the database (Fig.~\ref{fig:retrievalsspring}). The dataset built from a single 3.7~km walking route, captured with a Prophesee EVK4 synchronized to a GPS logger for ground truth annotation. The route was walked in a single direction of travel and captured in three camera orientations, front-on, left, and right, giving 11.1~km of database recordings in total. The queries are a separate set of routes that overlap the database at seven revisited areas, collected under two illumination conditions, morning and daytime; the results in Sect.~\ref{subsec:noveldata_results} combine the queries from both. In total, slicing event streams at 50~ms generated 132,569 database reference event frames, with 5,557 query event frames from 19 individual recordings. 

\subsection{Baseline methods}
\label{subsec:baselines}
We compare the recall performance of MegaEvent to \textbf{EventVLAD~\cite{Lee2021}}, \textbf{EventGeM~\cite{Hines2026}}, and \textbf{SpikeVPR~\cite{Keime2026}}. Event-VPR~\cite{Kong2022} could not be included as no code is publicly available. EventCV~\cite{eventcv} was used to generate event representations for each baseline: polarity histogram (EventVLAD, SpikeVPR), multi-channel time surface (EventGeM), and the representation for MegaEvent as described in~\cite{Cao2026}.

For SpikeVPR~\cite{Keime2026}, we followed the dataset-specific checkpoint selection described in the original implementation, avoiding checkpoints that overlapped with our evaluation data.

In addition, we compare against several state-of-the-art RGB VPR systems, as these have been shown to perform well on event frame data~\cite{Joseph2025}: MegaLoc~\cite{Berton2025}, SALAD~\cite{Izquierdo2024}, MixVPR~\cite{Ali2023}, CricaVPR~\cite{Lu2024}, BoQ~\cite{Ali2024}, SuperVLAD~\cite{lu2024supervlad}, and QAA~\cite{Xiao2025}.

\subsection{Localization metric}
\label{subsec:localmetrics}
We use Recall@1 (R@1) and Recall@10 (R@10) to measure localization accuracy~\cite{Schubert2024}: R@1 is the fraction of queries whose top-ranked reference is a correct match, whereas R@10 is the fraction for which a correct match appears anywhere in the top 10. We also report descriptor-to-match latency, defined as the wall-clock time for a single reference-query pair to go from raw events through event-frame construction, the backbone forward pass at $322\times322$~px, descriptor aggregation, and database retrieval. Latencies are reported for one such match rather than averaged over a batch.

\subsection{Training hyperparameters}
\label{subsec:trainhyper}
We fine-tuned our model with an AdamW optimizer with a learning rate of $5\times10^{-6}$ for the generative backbone~\cite{Cao2026} and $5\times10^{-5}$ for the SALAD~\cite{Izquierdo2024} aggregator, as the backbone is pre-trained and the SALAD layer is uninitialized. A weight decay of $0.001$ and a gradient clip of $1.0$ were implemented to control model overfitting. Training was initialized with a warm-up of $500$ steps, with a maximum of $2{,}000$ training steps, by which point the loss had converged and plateaued.

For ViT/S, we trained with $N=32$ places per stream and $4$ images per place, giving $128$ images per stream and an effective batch of $768$ images/step across the $6$ streams. Place identifiers are offset per stream, so a step covers $192$ distinct places, of which each loss term in Eq.~\eqref{eq:losstot} sees only the $32$ belonging to its own stream. For ViT/B, we trained with $N=64$ for an effective batch of $1536$ images/step. The multi-similarity loss was set with $\alpha=1.0$, $\beta=50.0$, and $\lambda=0.0$, and pairs were selected with a multi-similarity miner using a margin $\epsilon=0.1$.

To prevent over-fitting of the models from synthetic event streams generated from I2E~\cite{Ma2026}, we perturbed each event stream with three sensor non-idealities that the I2E simulator lacks. To mimic per-pixel variation in contrast thresholds, the event frame was scaled by a per-image gain $g \sim \mathcal{U}(0.7, 1.4)$, where $\mathcal{U}$ is a uniform distribution. The gain was applied after~Eq.~\eqref{eq:accumulate} because the percentile normalization is invariant to a rescaling of the raw counts. Next, to emulate events lost to refractory periods and dropouts, each pixel was independently reset to background with a probability $d \sim \mathcal{U}(0, 0.05)$ per frame. Finally, to simulate background activity and hot pixels, a fraction of pixels $s \sim \mathcal{U}(0, 0.02)$ received a spurious event of random polarity.

Our model follows MegaLoc~\cite{Berton2025} by including an additional linear projection after the SALAD~\cite{Izquierdo2024} aggregator to the final 8,448-dimensional descriptor. This includes an increase in the number of SALAD clusters from 64 to 256, as per~\cite{Berton2025}. We also train the ViT/S and ViT/B MegaEvent variants from the pre-trained backbones in~\cite{Cao2026}.

\subsection{Computational environment}
\label{subsec:compenv}
All models were trained on a high-performance computing cluster consisting of either A100 (40~GB VRAM) or H100 (80~GB VRAM) GPUs with 128~GB of memory and 16 CPU cores. All evaluations were performed on an RTX 2080 GPU (8~GB VRAM) with an 8-core CPU and 32~GB of memory.

        \begin{table*}[t]
        \footnotesize
        \centering
        \setlength{\tabcolsep}{5pt}

        \caption{Recall@1 (R@1) and Recall@10 (R@10) performance of event-based and frame-based VPR methods on real event VPR datasets. Column headings give the reference and query traverses as named by each dataset's authors. \textbf{Bold} is the best performing method, \underline{underline} is the second best performing; tied values receive the same emphasis.}

        \begin{tabular*}{\textwidth}{@{\extracolsep{\fill}}l*{12}{c}@{}}
            \toprule

            & \multicolumn{6}{c}{\textbf{Brisbane-Event-VPR}~\cite{Fischer2020}}
            & \multicolumn{4}{c}{\textbf{NSAVP}~\cite{Carmichael2025}}
            & \multicolumn{2}{c}{\textbf{NYC-Event-VPR}~\cite{Pan2025}} \\
    
            \cmidrule(lr){2-7}
            \cmidrule(lr){8-11}
            \cmidrule(lr){12-13}

            \multicolumn{1}{c}{Ref:Query} & \multicolumn{2}{c}{Sunset1:Daytime}
            & \multicolumn{2}{c}{Sunset1:Morning}
            & \multicolumn{2}{c}{Sunset1:Sunrise}
            & \multicolumn{2}{c}{R0-FS0:R0-FA0}
            & \multicolumn{2}{c}{R0-RS0:R0-RA0}
            & \multicolumn{2}{c}{Random 10\% split}
            \\

            \cmidrule(lr){1-1}
            \cmidrule(lr){2-3}
            \cmidrule(lr){4-5}
            \cmidrule(lr){6-7}
            \cmidrule(lr){8-9}
            \cmidrule(lr){10-11}
            \cmidrule(lr){12-13}

            Method
            & R@1 & R@10
            & R@1 & R@10
            & R@1 & R@10
            & R@1 & R@10
            & R@1 & R@10
            & R@1 & R@10 \\

            \midrule

            EventVLAD~\cite{Lee2021}
            & 0.06 & 0.19
            & 0.11 & 0.25
            & 0.19 & 0.38
            & 0.20 & 0.35
            & 0.25 & 0.40
            & 0.11 & 0.26 \\

            EventGeM~\cite{Hines2026}
            & 0.44 & 0.52
            & 0.71 & 0.76
            & 0.78 & 0.81
            & 0.54 & 0.57
            & 0.62 & 0.65
            & 0.65 & 0.80 \\

            SpikeVPR~\cite{Keime2026}
            & 0.14 & 0.36
            & 0.21 & 0.47
            & 0.26 & 0.55
            & 0.34 & 0.59
            & 0.48 & 0.70
            & 0.38 & 0.61 \\

            \midrule

            MegaLoc~\cite{Berton2025}
            & 0.52 & 0.72
            & 0.41 & 0.58
            & 0.62 & 0.76
            & 0.79 & 0.86
            & 0.74 & 0.79
            & 0.75 & 0.91 \\

            SALAD~\cite{Izquierdo2024}
            & 0.39 & 0.63
            & 0.38 & 0.57
            & 0.55 & 0.73
            & 0.75 & 0.84
            & 0.72 & 0.78
            & 0.68 & 0.86 \\

            MixVPR~\cite{Ali2023}
            & 0.55 & 0.75
            & 0.65 & 0.82
            & 0.83 & 0.92
            & 0.76 & 0.83
            & 0.74 & 0.79
            & 0.79 & 0.93 \\

            CricaVPR~\cite{Lu2024}
            & 0.45 & 0.69
            & 0.35 & 0.52
            & 0.58 & 0.74
            & 0.68 & 0.79
            & 0.70 & 0.75
            & 0.71 & 0.90 \\

            BoQ~\cite{Ali2024}
            & 0.55 & 0.76
            & 0.50 & 0.68
            & 0.67 & 0.81
            & 0.81 & \underline{0.87}
            & 0.76 & \underline{0.81}
            & 0.75 & 0.90 \\

            SuperVLAD~\cite{lu2024supervlad}
            & 0.44 & 0.67
            & 0.40 & 0.58
            & 0.59 & 0.76
            & 0.78 & 0.85
            & 0.71 & 0.77
            & 0.70 & 0.88 \\

            QAA~\cite{Xiao2025}
            & 0.58 & 0.80
            & 0.50 & 0.66
            & 0.71 & 0.82
            & \underline{0.82} & \textbf{0.88}
            & 0.74 & 0.79
            & 0.75 & 0.91 \\

            \midrule

            Ours (ViT/S)
            & \underline{0.66} & \underline{0.85}
            & \underline{0.73} & \underline{0.91}
            & \underline{0.85} & \underline{0.94}
            & 0.80 & 0.85
            & \underline{0.78} & \underline{0.81}
            & \underline{0.89} & \underline{0.97} \\

            Ours (ViT/B)
            & \textbf{0.71} & \textbf{0.87}
            & \textbf{0.78} & \textbf{0.93}
            & \textbf{0.88} & \textbf{0.95}
            & \textbf{0.84} & \textbf{0.88}
            & \textbf{0.81} & \textbf{0.84}
            & \textbf{0.91} & \textbf{0.98} \\

            \bottomrule
        \end{tabular*}
        \label{tab:recall_real}
    \end{table*}

\section{Results}
\label{sec:results}

    \begin{figure*}[!t]
        \centering
        \includegraphics[width=\textwidth]{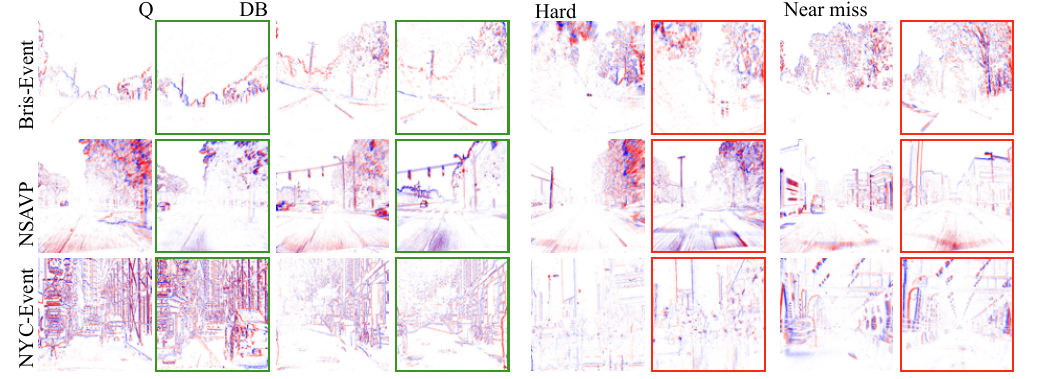}
        \caption{Top-1 retrievals from MegaEvent (ViT/B) on Brisbane-Event-VPR~\cite{Fischer2020} (top row), NSAVP~\cite{Carmichael2025} (middle row) and NYC-Event-VPR~\cite{Pan2025} (bottom row). Q: query; DB: retrieved database match. \textcolor{mygreen}{Green} indicates a correct retrieval and \textcolor{myred}{red} indicates an incorrect retrieval. Hard cases show severe aliasing or sparse event frames. Near misses are retrievals just outside the 25~m tolerance.}
        \vspace{-15pt}
        \label{fig:retrievals}
    \end{figure*}

In this section we evaluate MegaEvent on standard event-based VPR datasets (Sect.~\ref{subsec:localperf}) and on the newly contributed viewpoint and rotation variant Springfield-Event-VPR dataset (Sect.~\ref{subsec:noveldata_results}), analyze the effects of synthetic and real event streams on model performance (Sect.~\ref{subsec:real2sim}), and finally report the runtime performance of MegaEvent against baseline methods (Sect.~\ref{subsec:runtime}).

\subsection{Event-based VPR benchmark performance}
\label{subsec:localperf}
This experiment tests our first claim, that fine-tuning an event backbone on synthetically converted VPR data outperforms both existing event-based methods and frame-based models applied directly to event frames. Table~\ref{tab:recall_real} provides a summary of the main results. Our proposed MegaEvent (ViT/B) performs the best on every dataset, with a per-dataset average R@1 of 0.79 for Brisbane-Event-VPR~\cite{Fischer2020}, 0.83 for NSAVP~\cite{Carmichael2025}, and 0.91 for NYC-Event-VPR~\cite{Pan2025}. EventGeM~\cite{Hines2026} was the next best performing model on these datasets with average R@1 of 0.64, 0.58, and 0.65, respectively. We note that despite EventGeM having a two-stage matching protocol, it performed worse than our single-stage MegaEvent. Fig.~\ref{fig:retrievals} shows examples of retrieved places across the datasets.

In addition to the event-based models, we evaluated how standard RGB VPR methods, including MegaLoc~\cite{Berton2025}, perform on the event-based VPR datasets to identify an advantage of fine-tuning for the MegaEvent models (Table~\ref{tab:recall_real}). Averaged over the seven frame-based methods in Table~\ref{tab:recall_real}, MegaEvent (ViT/B) outperforms them by 0.26 R@1 on Brisbane-Event-VPR~\cite{Fischer2020}, 0.08 on NSAVP~\cite{Carmichael2025}, and 0.18 on NYC-Event-VPR~\cite{Pan2025}.

\begin{figure*}[!t]
        \centering
        \includegraphics[width=\textwidth]{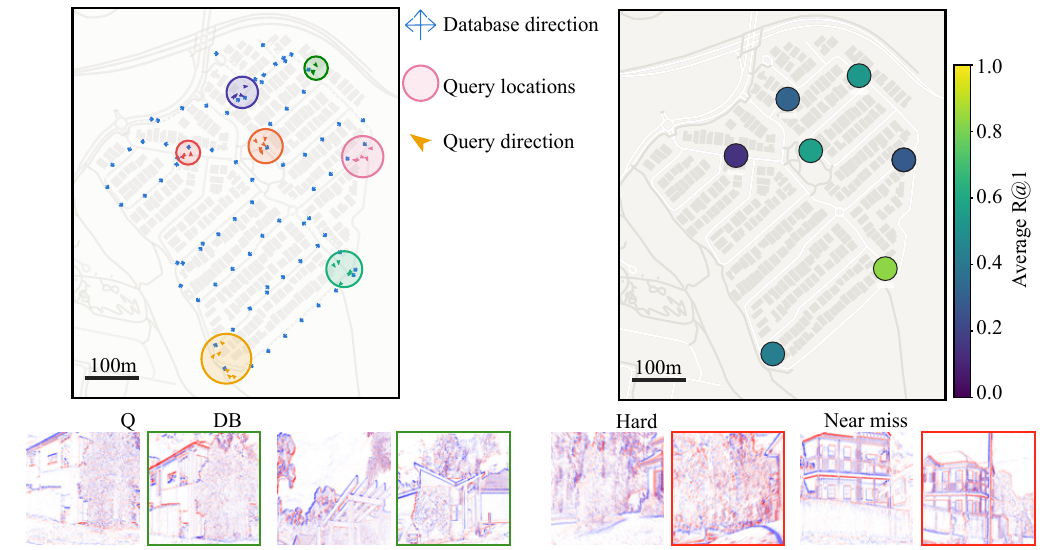}
        \caption{The Springfield-Event-VPR dataset. Left: database route and the seven query locations, with the legend shown alongside. Right: the same seven locations coloured by the average R@1 of MegaEvent (ViT/B) at each. Bottom: example top-1 retrievals across the dataset, with colour coding as in Fig.~\ref{fig:retrievals}.}
        \label{fig:retrievalsspring}
    \end{figure*}

\subsection{New viewpoint variant dataset}
\label{subsec:noveldata_results}
This experiment tests our central claim, that MegaEvent is robust to viewpoint change, using a dataset whose queries deliberately differ in heading from the database. The viewpoint variance for the query set was captured by revisiting the same location from different orientations and directions of travel (Fig.~\ref{fig:retrievalsspring}), matching to a reference database captured in one direction with 3 unique camera angles (Sect.~\ref{subsec:noveldata_setup}).

Table~\ref{tab:springfield} summarizes the results. Overall, MegaEvent ViT/B model performed the most robust localization with an R@1 of 0.50. The closest comparison is the frame-based MegaLoc~\cite{Berton2025}, which has previously been shown to perform accurate localization for event data~\cite{Joseph2025}, applied either zero-shot to the same event frames (R@1 of 0.39) or to intensity images reconstructed from the event stream with the recurrent E2VID model~\cite{Rebecq2019} (R@1 of 0.41). E2VID reconstruction~\cite{Rebecq2019} combined with MegaLoc outcompeted our ViT/S model. We did not isolate the cause; one possibility is MegaLoc's larger DINOv2~\cite{Oquab2024} backbone. Our MegaEvent ViT/B model outperformed the reconstructed frames with MegaLoc~\cite{Rebecq2019, Berton2025} by 0.09 R@1. The gain in recall between event frames and reconstructed intensity images for MegaLoc~\cite{Berton2025} is nonetheless marginal. This indicates that conventional VPR methods are compatible with event frame representations, and should remain a point of comparison~\cite{Joseph2025}. The best event-based baseline, EventGeM, achieved an R@1 of just 0.12.

    \begin{table}[!t]
        \footnotesize
        \centering

        \caption{Recall@1 (R@1) and Recall@10 (R@10) performance on the Springfield-Event-VPR dataset for MegaEvent and baseline methods. E2VID~\cite{Rebecq2019} denotes MegaLoc applied to intensity images reconstructed from the event stream. \textbf{Bold} is the best performing method, \underline{underline} is the second best performing.}

        \begin{tabular*}{\columnwidth}{@{\extracolsep{\fill}}l*{2}{c}@{}}
            \toprule

            & \multicolumn{2}{c}{\textbf{Springfield-Event-VPR}} \\

            \cmidrule(lr){2-3}

            Method & R@1 & R@10 \\

            \midrule

            EventVLAD~\cite{Lee2021} & 0.02 & 0.15  \\
            EventGeM~\cite{Hines2026} & 0.12 & 0.31   \\
            SpikeVPR~\cite{Keime2026} & 0.05 & 0.20  \\
            \midrule
            MegaLoc~\cite{Berton2025} & 0.39 & 0.54 \\
            MegaLoc~\cite{Berton2025} + E2VID~\cite{Rebecq2019} & \underline{0.41} & \underline{0.58} \\
            \midrule
            Ours (ViT/S) & 0.40 & 0.53  \\
            Ours (ViT/B) & \textbf{0.50} & \textbf{0.61}  \\

            \bottomrule

        \end{tabular*}
    \label{tab:springfield}
    \end{table}

This dataset, in addition to the benchmark VPR datasets in Sect.~\ref{subsec:localperf}, demonstrates the performance of MegaEvent on both conventional on-the-rails VPR datasets and challenging viewpoint variant tasks.

\subsection{Real2Sim analysis of synthetic and real event streams}
\label{subsec:real2sim}
This experiment asks whether MegaEvent's training on synthetic events leaves it advantaged on synthetic test data. As MegaEvent was trained on synthetic event streams, we investigated how our system performs on a VPR dataset converted to a synthetic event stream from RGB frames. Table~\ref{tab:real2sim} summarizes the results of converting RGB images from the Brisbane-Event-VPR dataset~\cite{Fischer2020} taken using the GoPro camera.

    \begin{table}[!t]
        \footnotesize
        \centering

        \setlength{\tabcolsep}{2pt}
        \caption{Comparison of GoPro RGB images converted to synthetic event streams using I2E~\cite{Ma2026} for Real2Sim evaluation from the Brisbane-Event-VPR dataset~\cite{Fischer2020} (Sunset1 reference, Morning query). \textcolor{myred}{Red} indicates a degradation in performance with synthetic events, \textcolor{mygreen}{green} indicates an improvement with synthetic events. EventGeM (global only) reports its global retrieval stage without RANSAC re-ranking.}

        \begin{tabular*}{\columnwidth}{@{\extracolsep{\fill}}l*{6}{c}@{}}
            \toprule

            & \multicolumn{2}{c}{Real events} & \multicolumn{2}{c}{Synthetic events} & \multicolumn{2}{c}{$\Delta$ Real2Sim} \\

            \cmidrule(lr){2-3}\cmidrule(lr){4-5}\cmidrule(lr){6-7}
            Method & R@1 & R@10 & R@1 & R@10 & R@1 & R@10\\

            \midrule

            EventVLAD~\cite{Lee2021} & 0.11 & 0.25 & 0.22 & 0.40 & \color{mygreen}0.11 & \color{mygreen}0.15 \\
            EventGeM (global only)~\cite{Hines2026} & 0.40 & 0.45 & 0.26 & 0.31 & \color{myred}-0.14 & \color{myred}-0.14 \\
            EventGeM~\cite{Hines2026} & 0.71 & 0.76 & 0.57 & 0.65 & \color{myred}-0.14 & \color{myred}-0.11 \\
            SpikeVPR~\cite{Keime2026} & 0.21 & 0.47 & 0.25 & 0.52 & \color{mygreen}0.04 & \color{mygreen}0.05 \\
            \midrule
            Ours (ViT/S) & 0.73 & 0.91 & 0.81 & 0.94 & \color{mygreen}0.08 & \color{mygreen}0.03 \\
            Ours (ViT/B) & 0.78 & 0.93 & 0.84 & 0.94 & \color{mygreen}0.06 & \color{mygreen}0.01 \\

            \bottomrule

        \end{tabular*}
    \label{tab:real2sim}
    \end{table}

Each of the methods, except for EventGeM, experienced an increase in recall performance using synthetic events compared to real event streams. Overall, MegaEvent did not gain a substantial domain advantage over other methods when using synthetic event streams. EventVLAD~\cite{Lee2021} gained the largest benefit of an increase of 0.11 absolute R@1. For EventGeM~\cite{Hines2026}, the two-stage re-rank component provides a +0.31 boost to global R@1 for both real and synthetic events. The degradation for synthetic events therefore originates in the global retrieval stage rather than in re-ranking. EventGeM is also the only method that degrades, which is consistent with its local keypoint matching depending on structure that a single converted frame does not reproduce as reliably as a 50~ms window of real events.

\subsection{Runtime performance}
\label{subsec:runtime}
This experiment establishes the runtime cost of the recall reported above, so that the ViT/S and ViT/B variants can be weighed against each other and against the event-based baselines on a like-for-like basis. We compare the runtime of baseline localization systems using descriptor-to-match latency, alongside parameter count and average recall in Table~\ref{tab:modelparams}. Our ViT/S model achieved an average R@1 within 0.03 of ViT/B while reducing descriptor-to-match latency from 22.3~ms to 19.4~ms. EventGeM~\cite{Hines2026} has a relatively low parameter count due to its compact backbone, but its overall latency is 30.3~ms because of the computationally expensive RANSAC re-ranking step. EventVLAD~\cite{Lee2021} performed worst overall, with the highest parameter count among the event-based baselines, the slowest latency, and the lowest average recall. SpikeVPR~\cite{Keime2026} had the lowest parameter count and latency (12.3~ms), but its recall is substantially below that of MegaEvent and EventGeM.

    \begin{table}[!t]
        \footnotesize
        \centering

        \setlength{\tabcolsep}{2pt}
        \caption{Model size (M parameters), descriptor-to-match latency (ms, defined in Sect.~\ref{subsec:localmetrics}), global descriptor dimension, and average R@1 (the mean over the six reference-query pairs in Table~\ref{tab:recall_real}). $\uparrow$ indicates higher values are better, $\downarrow$ indicates lower values are better. \textbf{Bold} is the best performing method, \underline{underline} is the second best performing.}

        \begin{tabular*}{\columnwidth}{@{\extracolsep{\fill}}l*{4}{c}@{}}
            \toprule

            & \multicolumn{1}{c}{Parameters $\downarrow$} & \multicolumn{1}{c}{Latency $\downarrow$} & \multicolumn{1}{c}{Descriptor} & \multicolumn{1}{c}{R@1 $\uparrow$} \\

            Method & (M) & (ms) & (dim.) & (Average) \\

            \midrule

            EventVLAD~\cite{Lee2021} & 175.50 & 47.73 & 1000 & 0.15 \\
            EventGeM~\cite{Hines2026} & \underline{23.5} & 30.25 & 384 & 0.62 \\
            SpikeVPR~\cite{Keime2026} & \textbf{2.92} & \textbf{12.25} & 4096 & 0.30 \\
            \midrule
            Ours (ViT/S) & 163.5 & 19.41 & 8448 & \underline{0.79} \\
            Ours (ViT/B) & 228.6 & 22.34 & 8448 & \textbf{0.82} \\

            \bottomrule

        \end{tabular*}
    \label{tab:modelparams}
    \end{table}

\section{Discussion and Future Directions}
\label{sec:discuss}
We present MegaEvent, an event-based VPR system capable of accurately localizing under viewpoint variance. We showed that a model trained entirely on synthetically converted event streams achieves state-of-the-art recall on real event data, and that its advantage does not come from a synthetic-domain bias.

MegaEvent extends event-based localization enabling multi-viewpoint robustness, with an 8 to 26 point R@1 margin over frame-based VPR models applied directly to event frames and a larger margin over existing event-based methods. We demonstrate that VPR models trained on synthetic event streams transfer effectively to real event streams across three sensor types, a DAVIS346 (Brisbane-Event-VPR~\cite{Fischer2020}), a DVXplorer (NSAVP~\cite{Carmichael2025}) and a Prophesee EVK4 (NYC-Event-VPR~\cite{Pan2025} and our Springfield-Event-VPR dataset), whilst avoiding over-fitting to a synthetic event domain. This generalizability across sensor types and localization scenarios (driving, walking) establishes MegaEvent as the strongest event-based VPR method on the benchmarks tested. Of particular interest is how well standard RGB VPR methods perform on event data, which has previously been shown~\cite{Joseph2025}, as event frames retain enough scene structure for these models to extract and match features across the reference and query sets.

We do note certain limitations in the implementation, however. MegaEvent is trained entirely on synthetically converted events, so no real-event training data constrains the descriptor space, and illumination invariance remains unresolved. Every training and evaluation stream used here was captured without camera bias adjustment, so severe lighting shifts are not represented. Future work will therefore explore how to improve illumination invariance for event cameras. We also plan to capture large-scale geo-localization datasets with event cameras, so that training on synthetic and real event streams can be compared directly, and so that a hybrid training regime that further closes the Sim2Real gap can be developed.

\bibliography{Bibliography}
\end{document}